\documentclass[runningheads]{llncs}
\usepackage{scalerel,graphicx,amssymb,amsmath,mathtools,wasysym,floatrow}
\usepackage[outercaption]{sidecap} 
\usepackage[ruled,vlined,linesnumbered]{algorithm2e}
\usepackage[dvipsnames]{xcolor}
\usepackage{txfonts}
\usepackage{tikz}
\usepackage{multirow}
\usetikzlibrary{arrows}
\usepackage{wrapfig}
\usepackage{todonotes}
\usepackage[inline]{enumitem}
\usepackage{array} 
\newcolumntype{M}[1]{>{\raggedright\arraybackslash}m{#1cm}}
\newcolumntype{N}[1]{>{\centering\arraybackslash}m{#1cm}}

\DeclareMathSymbol{\mlq}{\mathord}{operators}{``}
\DeclareMathSymbol{\mrq}{\mathord}{operators}{`'}

\newcommand\SF{\boldsymbol{SF}}

\newcommand\eq{\mathcal{EQ}}

\newcommand\sitf{\mathit{sf}}

\newcommand\asf{\boldsymbol{\mathit{ASF}}}

\newcommand{\car}{\ensuremath{%
		\bullet\kern-5pt
		\raise1pt\hbox{$\mathord{\rightarrow}$}}}
\def\cra{\hbox{$\multimapdotinv$}\kern-7.5pt\hbox{$\rightarrow$}}

\begin{document}
	\title{Case Level Counterfactual Reasoning in Process Mining}
	\author{Mahnaz Sadat Qafari \ \ \ 
		Wil van der Aalst}
	\authorrunning{M. S. Qafari et al.}
	%
	\institute{Rheinisch-Westfälische Technische Hochschule Aachen (RWTH), Aachen, Germany \\
		\email{m.s.qafari@pads.rwth-aachen.de,wvdaalst@pads.rwth-aachen.de}}
	
	\maketitle 
	\begin{abstract}
    Process mining is widely used to diagnose processes and uncover performance and compliance problems. It is also possible to see relations between different behavioral aspects, e.g., cases that deviate more at the beginning of the process tend to get delayed in the later part of the process. However, correlations do not necessarily reveal causalities.  Moreover, standard process mining diagnostics do not indicate how to improve the process. This is the reason we advocate the use of \emph{structural equation models} and \emph{counterfactual reasoning}. We use results from causal inference and adapt these to be able to reason over event logs and process interventions. We have implemented the approach as a ProM plug-in and have evaluated it on several data sets.
		
	\end{abstract}
\section{Introduction}
    Humans tend to learn from the past (their experiences) by analyzing possible alternatives of what happened in the reality and reflecting on their findings aiming for better results in future similar cases (e.g., not doing the same mistakes). Thinking about possible alternatives to what happened in the past is called \emph{counterfactual thinking}. 
	
	The information systems of companies save data about the process instances (cases) in their event logs. Process mining extracts knowledge from the event logs for discovering the process model, monitoring process KPIs, and improving processes. Process improvement requires a deep comprehension of the process behavior and its cases. In this paper, we tailor the concept of \emph{counterfactual thinking} to process mining and explain why a specific situation has a special outcome. Given an instance with an undesirable outcome, we aim at providing a set of counterfactual statements (we call them \emph{explanations}) to explain such an outcome. 
	
    Companies can boost customer satisfaction and build trust by providing explanations for their specific cases without putting other people's rights and privacy in danger \cite{reddix2011credit}. Case level explanation can be used to explain why a customer has received a particular result, was it fair, or how to approach to get a better result. Moreover, the process manager can benefit from this method as it can be used to explain why something happens in a specific case and how to act differently to get different results in the future.
	
    Two important aspects of an explanation are accuracy and applicability. Both of them can be amended by distinguishing between correlation and causation among the process features, which prevents misleading explanations that recommend altering features with non-causal relationships with the result. For this matter, we propose using the \emph{structural equation model (SEM)} of the features in the procedure of generating explanations.
    
    Case level explanations are case dependent, which means, an explanation that is useful for a customer may not be favored by another customer with the same undesirable outcome. To overcome this issue, in the proposed method, we present a set of diverse explanations (i.e. explanations that differ from the given instance in different features) to the user such that the user can decide which one to apply.	Moreover, as the explanations are meant to be used by the human, the readability and understandability of the explanations are important. Therefore those explanations with a smaller number of features with different values from the given instance, are preferred \cite{wachter2017counterfactual}.

    The rest of the paper is organized as follows. In Section \ref{sec::relatedWork}, a brief overview of the related work is presented. In Section \ref{sec::method}, the method is presented. The experimental results are presented in Section \ref{sec::results}. Finally, in Section \ref{sec::conclusion} the conclusion is presented.
    \begin{figure}[t]
    \centering
    \includegraphics[width=120mm]{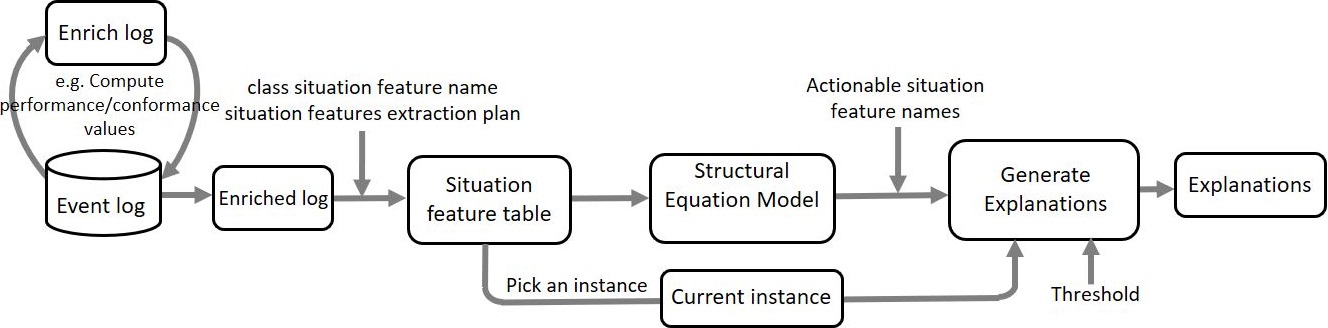}
    \caption{The general overview of the proposed method.}
    \label{fig::generalView}
\end{figure}
\section{Related Work} \label{sec::relatedWork}
There are already several approaches in the domain of process mining that deal with root cause analysis using the findings of a classification techniques \cite{suriadi2012root,ferreira2015using}. The drawback of these methods is that the classification techniques are based on correlation and not causal relationships. Also, there are several works considering causal relationships among different process features at the process level \cite{hompes2017discovering,narendar2019,me:)}. Moreover, in \cite{marlon} a method for generating case-level
recommendations of treatments that maximize the probability
of a given outcome is proposed. In this method a subset of candidate treatments that are most correlated with the outcome is extracted by applying an association rule mining technique. Then the subgroups with causal relation between treatment and outcome are identified using uplift tree. Finally, the subgroups are sorted by the ratio of the score associated to them by the uplift trees and their cost.

It is worth noting that counterfactual reasoning for explainability has been studied extensively in the field of data mining and machine learning (e.g., \cite{russell2019efficient,wachter2017counterfactual}).
\section{Method}
\label{sec::method}
The general overview of the proposed method is presented in Figure. \ref{fig::generalView}. First, we enrich the event log. Then, several random counterfactual instances similar to the current instance are generated. Among them, those that have a desirable outcome regarding a given threshold are selected and optimization techniques are used to make them as close as possible to the current instance. The resulting desirable counterfactual instances are ordered according to their distance with the current instance, and finally, converted into a set of explanations and presented to the people involved.

In the following, we first explain how we extract the data from the event log and then we describe the explanation generation method.
\vspace{-2mm}
\subsection{Situation Feature Table Extraction}
\vspace{-1mm}
Here, we mention how to extract features from the event logs of processes. An event log is a collection of traces where each trace is a collection of events. Each event indicates that specific activity has happened at a specific time for a specific case. A trace is a sequence of chronologically ordered events that belong to a specific case. Both traces and events may have several attributes. Here we assume the uniqueness of the events in the event log. We define an event log as follows:
\begin{definition}[Event Log]
An event log is a set of traces where each trace is composed of a chronologically ordered sequence of events and trace attributes (if applies). Moreover, each event
refers to a case, an activity, a timestamp, and event attributes (if applicable).
\end{definition}
\begin{figure}[t]
    \centering
    \includegraphics[width=80mm]{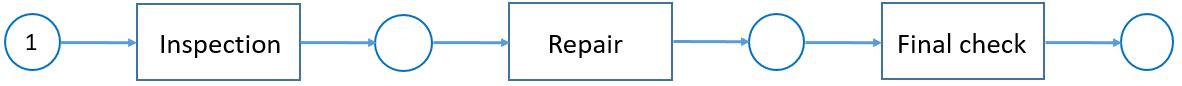}
    \caption{The process of the repair company.}
    \label{fig::process}
\end{figure}

Through this paper, we consider a company that repairs a specific product as the running example. The Petri-net model of this repair company is shown in Figure \ref{fig::process}. Each trace in the event log is corresponding to the process of repairing one product. In the ``inspection" activity, several tests are done to determine the defects of a product. Then it is repaired, and afterward ``final check" is done. We know that the newer products are more complicated and harder to deal with. In this example, ``model" is a trace level attribute where newer products have higher model numbers. ``team size" is another trace-level attribute that indicates the number of resources involved in repairing the product. ``num test" is an event-level attribute indicating the number of tests that have been done in the ``inspection" activity. A snapshot of an event log is shown in Table \ref{table:eventLog}. The manager of the repair company believes that in the trace with ``case id"= $c1$ the ``repair" activity (event $e2$) was too long and should have taken at most 500 hours. He/She needs to know if it was the case, and if so, how they could had prevented it.
\begin{table}[b]
    \centering
    \resizebox{0.6\textwidth}{!}{
    \begin{tabular}{|c|c|c|c|c|c|c|}
    \hline
        event id & case id & activity name & timestamp & team size & num test & model\\
        \hline
        e1 & c1 & inspection & 01-04-2020T08:00:00 & 2 & 42 & 7 \\
        e2 & c1 & repair & 04-04-2020T07:00:00 & 2 & 42& 7 \\
        e3 & c1 & final test & 28-04-2020T08:00:00 & 2 & 42 & 7 \\
        e4 & c2 & inspection & 01-05-2020T08:00:00 & 3 & 26 & 5 \\
        e5 & c2 & repair & 03-05-2020T11:00:00 & 3 & 26 & 5 \\
        e6 & c2 & final test & 19-05-2020T20:00:00 & 3 & 26 & 5 \\
        $\vdots$ & $\vdots$ & $\vdots$ & $\vdots$ & $\vdots$ & $\vdots$ & $\vdots$ \\
        \hline
    \end{tabular}
    }
    \caption{A snapshot of the event log of the repair company.}
    \label{table:eventLog}
\end{table}

When we are interested in features not existing in the event log, we need to enrich the event log by adding new derived features from the event log or possibly other sources to its traces and events. For example, we can enrich the repair company event log by adding an attribute named ``duration" to its events indicating the duration of that event in hours. In the repair company example, the value of the ``duration" attribute can be computed by subtracting the timestamp of two consecutive events in each trace.


With respect to the time precedence of the cause and effect, we consider just the features that have been extracted from that part of a trace that has been recorded before a specific feature as possible causes of it. For example, in the repair company, extracting the data from the ``final check" activity is meaningless when we want to find the features that causally influence the ``duration" of the ``repair" activity. So we need to extract the data from a proper prefix of a trace, which we call a \emph{situation}. Also, we define the \emph{situation set} of an event log as the set of all situations generated using its traces. Some of the subsets of the situation set of a given event log are more meaningful. For example the set of all the situations that end with an event with a specific activity name or the set of all traces. In the repair company example, we can have a subset of situations that end with an event whose activity name is ``repair". In this case, the situation subset would include all the prefixes of traces which include the events with the activity name ``inspection" and ``repair". The situation extracted from the first trace would include the two events with ``event id" $e1$ and $e2$. Let's call this situation $s_1$.

An event log may have both event and trace level attributes. Moreover, it is possible to have the same attribute name in both levels. To concretely determine the attributes that we are interested in their values, we use \emph{situation feature} notion. A situation feature refers to an attribute and an activity name (or possibly ``trace"). For example in the repair company, $\sitf_{team size}$ and $\sitf_{model}$ are two situation features indicating ``team size" and ``model" attributes in the trace level. While, $\sitf_{inspDuration}$ and $\sitf_{inspNumTest}$ are the situation features referring to the ``duration" and ``num test" in the ``inspection" activity. Also, $\sitf_{repairDuration}$ refers to the ``duration" of the ``repair" activity. The situation feature value extraction mechanism from a given situation is as follows:
\begin{itemize}
    \item If the situation feature refers to an attribute name and ``trace", then the value of that attribute in the trace-level is assigned to the situation feature.
    \item If the situation feature refers to an attribute name and an activity name, then the value of that attribute from an event with the given activity name, with the maximum timestamp is assigned to the situation feature.
\end{itemize}

For example, for the situation $s_1$, the value assigned to $\sitf_{inspDuration}$ is 71 (computed using timestamps) and the value assigned to $\sitf_{model}$ is 7.

To generate explanations, we need to know the situation feature that identifies the problem (we call it \emph{target situation feature}) and a set of descriptive situation features that are those features that may have causal effect on the problem. We call the set including the descriptive situation features and the target situation feature a \emph{situation feature extraction plan} and denote it by $\SF$. We can look at the $\SF$ as the schema in a tabular data. For example in the repair company, as the manager believes that the duration of ``repair" activity for some cases should have been shorter, the target situation feature is $\sitf_{repairDuration}$. Also he has considered $\sitf_{model}$, $\sitf_{team size}$, $\sitf_{inspNumTest}$, and $\sitf_{inspDuration}$ as descriptive situation features. So, in this example we have $\SF_{repair} =\{\sitf_{model},\sitf_{team size}, \sitf_{inspNumTest},\sitf_{inspDuration},\sitf_{repairDuration}\}$.

Given a situation feature extraction plan, $\SF$ we can map each situation to a data point by simply extracting the values of situation features in $\SF$ using the proper mechanism. We call such a data point an \emph{instance}. Moreover, we can define a target-dependent tabular data, called \emph{situation feature table}, extracted from a given situation subset, as the bag of the instances extracted from the situations in a given situation subset. As an example, using $\SF_{repair}$ instance $i_{repair}=\{(\sitf_{model},7),(\sitf_{team size},2), (\sitf_{inspNumTest},42),\\(\sitf_{inspDuration},71),(\sitf_{repairDuration},577)\}$ is generated from $s_1$.

\subsection{Explanation Generation Method}

Consider an instance $i$ in the situation feature table with an undesirable target situation feature value regarding a threshold $t$. For example, in the repair company the threshold is 500. W.l.o.g., in this paper, we always assume that the values lower than the given threshold are desirable. Explanations are diverse instances which are close to $i$ and have a desirable target situation feature value. As it is vain to investigate the effect of intervention on those situation features that their value can not be altered by the user, we study the effect of changing the value of those situation features that are modifiable by the user. We call the set of modifiable situation features \emph{actionable situation features} and denote it with $\asf$. We define a set of counterfactual explanations for a given instances as follows. 
\begin{definition}[A Set of Counterfactual Explanation]
Let $i$ be an instance for which the target situation feature value is undesirable. A set of explanations for $i$ is a set of diverse instances that are close to $i$ and yet differ from $i$ in a subset of $\asf$ and have a desirable result for the target situation feature.
\end{definition}

To generate the set of counterfactual explanations, we take the following three steps: 

\textbf{1. Generating candidates.} We generate several candidates for the values that could had been assigned to the actionable situation features. Each candidate is a value assignment to a subset of situation features in $\asf$. We generate candidates such that for half of them the situation feature values are selected from their distribution in the situation feature table and for the other half, they are selected randomly from their domain. 

 \textbf{2. Predicting the value of the target situation feature. }In the second step, we compute the effect of replacing the values of the situation features in the given instance with those in the generated candidates on the value of target situation feature using the SEM of the situation features. The SEM of the situation features of a situation feature table can be provided by a customer who possesses the process domain knowledge or can be inferred in a data-driven manner using several methods that already exist in the literature (e.g., \cite{me:),narendar2019}). Loosely speaking, a SEM is a set of equations that determine how to generate the observational and interventional distributions. More formally:
\begin{definition}[Structural Equation Model (SEM)]\label{SEM}
    Let $\SF$ be a situation feature extraction plan, the SEM of $\SF$ is defined as $\eq \in \SF \to Expr(\SF)$ where for each $
    \sitf \in SF$, $Expr (\SF)$ is an expression over the situation features in $\SF$ and possibly some noise $N_{\sitf}$. Moreover, the noise distributions of $N_{\sitf}$ for all $\sitf \in \SF$ have to be mutually independent.
\end{definition}
 We assume that $\SF$ includes all relevant situation features and there is no common hidden confounder for the situation features in $\SF$. Also, we assume that the SEM does not include any loop. In Table \ref{tab::SEM}, a possible SEM for the repair company is presented.
 
\begin{table}[b]
	\begin{center}
		\caption{A possible SEM for the repair company.}
		\resizebox{0.8\textwidth}{!}{			
			\begin{tabular}{l l}
				$\sitf_{model} = N_{\sitf_{model}}$  & $N_{\sitf_{model}} \sim Uniform(1,10)$  \\
			    $\sitf_{team size} = N_{\sitf_{team size}}$ &$N_{\sitf_{team size}}\sim Uniform (1,3)$ \\
			    $\sitf_{repairDuration} =10 \sitf_{model} + N_{\sitf_{repairDuration}}$ & $N_{\sitf_{repairDuration}} \sim Uniform(-2,4)$  \\
		    	$\sitf_{inspNumTest} =5\sitf_{model} + 3\sitf_{team size}+N_{\sitf_{inspNumTest}}$ &   $N_{\sitf_{inspNumTest}} \sim Uniform(-1,2)$ \\
			    $\sitf_{repairDuration} =50\sitf_{model} + 5\sitf_{inspNumTest} + N_{\sitf_{repairDuration}}$  &$N_{\sitf_{repairDuration}} \sim Uniform(10,20)$  \\
		\end{tabular}}
		\label{tab::SEM}
	\end{center}
\end{table}

Using SEM $\eq$, prediction of the class situation feature value for each candidate involves three steps \emph{abduction}, \emph{action}, and \emph{prediction} \cite{pearl2000models}. We explain these steps using the repair company example.
    \begin{itemize}
    \vspace{-1mm}
        \item \textbf{Abduction. }First we need to incorporate the observed data, instance $i$, into the model, $\eq$, and generate a \emph{counterfactual SEM} that explains the conditions and the behavior of the system and the environment when $i$ was happening. A \emph{counterfactual SEM}, $\eq'$, is obtained by replacing the distribution of noise terms in $\eq$ with the corresponding noise distributions condition on $\SF = i$. 
         Considering the SEM in Table \ref{tab::SEM} and $i_{repair}$, the equations of the counterfactual SEM $\eq_{repair}'$ are: $\sitf_{model} = 7$, $\sitf_{inspNumTest} = 2$,  $\sitf_{inspDuration} =10 \sitf_{model} +1$, $\sitf_{inspNumTest} =~5 \sitf_{model} + 3\sitf_{teamSize}+1$, and  $\sitf_{repairDuration} =50\sitf_{model} + 5\sitf_{inspNumTest} +~17$.

        \item \textbf{Action. }The second step is taking action toward enforcing changes in the counterfactual SEM $\eq'$, regarding candidate $c$. The result is a SEM $\eq''$ where $\sitf = c_{\sitf}$ where $c_{\sitf}$ is the value assigned to $\sitf$ by $c$ if $\sitf \in dom(c)$ and $\sitf = \eq'(\sitf)$ where $\sitf$ is not in the domain of $c$. As an example,
        suppose that we are interested in predicting the value of $\sitf_{repairDuration}$ for the candidate $\{(\sitf_{teamSize},3)\}$. Intervention on the counterfactual SEM $\eq_{repair}'$, results in replacing $\sitf_{teamSize} = 2$ with $\sitf_{teamSize} =~3$.
        \item \textbf{Prediction. }The third step involves using the modified SEM to predict the counterfactual value of the target situation feature by simply computing the value of targer situation feature (or its distribution) in the counterfactual SEM under the intervention. In this step, we remove those situation features from the domain of $c$ that do not affect the target situation feature value. In the above example, computing the values of the situation features we have:
        $\{((\sitf_{model}, \bot),7), (\sitf_{teamSize}, 3),(\sitf_{inspNumTest},\\45), (\sitf_{inspDuration},71), (\sitf_{repairDuration},592)\}$. We call such an instance a \emph{counterfactual instance}.
    \end{itemize}

 \textbf{3. Selecting a subset of candidates.} We want explanations to be a set of diverse candidates with a small domain and a desirable predicted target situation feature value. Also we want them to be close to the given instance. To compute the distance between instances, we use $L_1$ metric on the normalized situation features. As mentioned in \cite{wachter2017counterfactual}, using $L_1$ metric, more sparse explanations would be generated. For the diversity, we partition candidates with desirable predicted outcome based on their domain and then sort them in each partition according to their distance from the given instance. A set of these candidates are selected one by one from different partitions, with the priority of those partitions that have a smaller domain.


\section{Experimental Results} \label{sec::results}
The implemented plugin is available in ProM nightly build under the name \emph{counterfactual explanation}. In the implemented plugin, we can apply several classifiers (including Regression Tree (RT), Locally Weighted Learning (LWL), Multi-layer perceptron (NN)), as well as SEM, to predict the target situation feature value of candidates.

We applied the implemented plugin on a synthetic event log to see how different might be the explanations generated by the SEM and by a machine learning technique with the highest accuracy in terms of predicted target situation feature values and the number of situation features with different values in the given instance and the explanations. So, we did not use optimization on the selected desirable counterfactual instances.

For the synthetic event log, we have used the repair company example and $i_{repair}$ as the instance with the undesirable target situation feature. Also, the values lower than the given threshold 500 were desirable. We considered all the descriptive situation features as actionable. We have generated 1000 traces such that the SEM of its situation feature values is the one in Table \ref{tab::SEM}. Then, we generate a set of 8 explanations by generating several candidates and using the SEM in Tabble \ref{tab::SEM} to evaluate them.

We have used the classifier with the highest accuracy for predicting the value of $\sitf_{rapairDuration}$ on the selected candidates in the previous step. The accuracy of RT, LWL, and NN on the data were 0.818, 0.990, and 0.984, respectively. But their accuracy reduced on the counterfactual instances to 0.74, 0.77, and 0.76, respectively.

\begin{figure}[t]
    \centering
    \includegraphics[width=110mm]{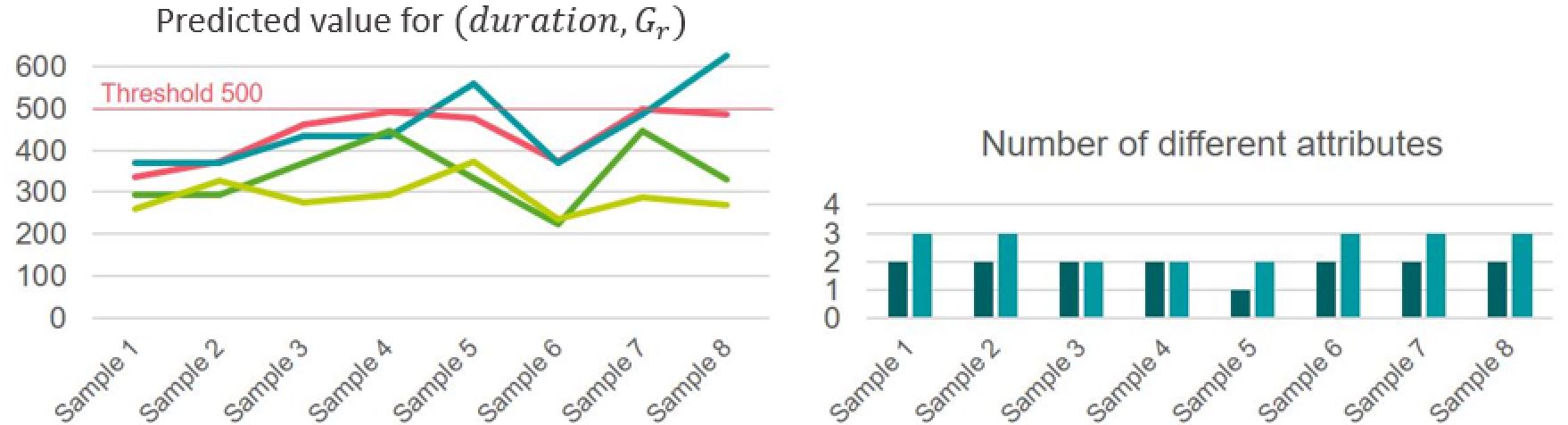}
    \caption{The result of applying the implemented method on the synthetic event logs.}
    \label{fig::resLinear}
\end{figure}

The results of applying the proposed method using SEM and three mentioned machine learning techniques are presented in Figure \ref{fig::resLinear}. In left part of Figure \ref{fig::resLinear}, the predicted $\sitf_{rapairDuration}$ of the selected desirable candidates using SEM (red line), RT (blue line), LWL (green line), and NN (light green line) are presented. In the right side of Figure \ref{fig::resLinear}, the size of the domain of the selected candidates is demonstrated.
\paragraph{\textbf{Discussion.}} \label{sec::discussion}
As demonstrated in Figures \ref{fig::resLinear}, there is a gap between the values predicted by the machine learning techniques and by SEM. Also, the accuracy of the classifiers predicting the value of the counterfactual instances drops dramatically. This phenomenon can be explained by the difference in their mechanism of predicting counterfactual values. Using a machine learning technique, neither the behavior of the environment nor the effect of an intervention is considered; but, the generated instance is regarded as a new instance, which may result in wrong predictions.

The difference in the number of effective situation features with different values between the given and explanations comes from the fact that machine learning techniques do not distinguish among the situation features with causal and mere correlation relationship with the target situation feature. On the other hand, using SEM the changes in the values of the situation features that have no causal relationships with the target situation feature in the counterfactual instances are simply ignored.

\section{Conclusion} \label{sec::conclusion}
We have presented a method that can be used by companies to explain to their customers why they have received a specific outcome in a case-specific manner and help them to prevent the same outcome in the future. As a result, the interpretability and accountability of the companies would be boosted.

The results of the evaluations have shown that ignoring the causal relationships among the situation features may end up in explanations that suggest changing situation features with no causal effect on the class situation feature. Moreover, using a machine learning technique, regardless of its accuracy, for predicting the value of the target situation feature may result in wrong explanations or missing some of the good explanations. 
\bibliographystyle{splncs04}
 \bibliography{biblio}

\begin{thebibliography}{10}
\providecommand{\url}[1]{\texttt{#1}}
\providecommand{\urlprefix}{URL }
\providecommand{\doi}[1]{https://doi.org/#1}

\bibitem{marlon}
Bozorgi, Z.D., Teinemaa, I., Dumas, M., Rosa, M.L., Polyvyanyy, A.: Process
  mining meets causal machine learning:discovering causal rules from event
  logs. In: ICPM (2020)

\bibitem{ferreira2015using}
Ferreira, D.R., Vasilyev, E.: Using logical decision trees to discover the
  cause of process delays from event logs. Computers in Industry  \textbf{70},
  194--207 (2015)

\bibitem{hompes2017discovering}
Hompes, B.F., Maaradji, A., La~Rosa, M., Dumas, M., Buijs, J.C., van~der Aalst,
  W.M.: Discovering causal factors explaining business process performance
  variation. In: International Conference on Advanced Information Systems
  Engineering. pp. 177--192. Springer (2017)

\bibitem{narendar2019}
Narendra, T., Agarwal, P., Gupta, M., Dechu, S.: Counterfactual reasoning for
  process optimization using structural causal models. In: Proceedings of
  Business Process Management Forum. vol.~360, pp. 91--106. Springer (2019).
  \doi{10.1007/978-3-030-26643-1\_6}

\bibitem{pearl2000models}
Pearl, J., et~al.: Models, reasoning and inference. Cambridge, UK:
  CambridgeUniversityPress  (2000)

\bibitem{peters2017elements}
Peters, J., Janzing, D., Sch{\"o}lkopf, B.: Elements of causal inference:
  foundations and learning algorithms. MIT press (2017)

\bibitem{me:)}
Qafari, M.S., van~der Aalst, W.: Root cause analysis in process mining using
  structural equation models. In: BPI (2020)

\bibitem{reddix2011credit}
Reddix-Smalls, B.: Credit scoring and trade secrecy: An algorithmic quagmire or
  how the lack of transparency in complex financial models scuttled the finance
  market. UC Davis Bus. LJ  \textbf{12}, ~87 (2011)

\bibitem{russell2019efficient}
Russell, C.: Efficient search for diverse coherent explanations. In:
  Proceedings of the Conference on Fairness, Accountability, and Transparency.
  pp. 20--28 (2019)

\bibitem{suriadi2012root}
Suriadi, S., Ouyang, C., van~der Aalst, W.M., ter Hofstede, A.H.: Root cause
  analysis with enriched process logs. In: International Conference on Business
  Process Management. pp. 174--186. Springer (2012)

\bibitem{wachter2017counterfactual}
Wachter, S., Mittelstadt, B., Russell, C.: Counterfactual explanations without
  opening the black box: Automated decisions and the gdpr. Harv. JL \& Tech.
  \textbf{31}, ~841 (2017)

\end{thebibliography}
\end{document}